\pdfoutput=1
\documentclass[11pt]{article}
\usepackage{emnlp2021}
\usepackage{times}
\usepackage{latexsym}
\usepackage[T1]{fontenc}
\usepackage[utf8]{inputenc}
\usepackage{microtype}
\usepackage{graphicx}

\title{JuriBERT: A Masked-Language Model Adaptation for French Legal Text}

\author{Stella Douka \\
  \'Ecole Polytechnique \\
  AUEB \\
  \And
  Hadi Abdine \\
  \'Ecole Polytechnique \\
  \And
  Michalis Vazirgiannis \\
  \'Ecole Polytechnique \\
  AUEB \\ 
 \AND
 Rajaa El Hamdani \\
 HEC Paris \\
 \And
 David Restrepo Amariles \\
 HEC Paris \\
 }

\begin{document}

\maketitle
\begin{abstract}
Language models have proven to be very useful when adapted to specific domains. Nonetheless, little research has been done on the adaptation of domain-specific BERT models in the French language. In this paper, we focus on creating a language model adapted to French legal text with the goal of helping law professionals. We conclude that some specific tasks do not benefit from generic language models pre-trained on large amounts of data. We explore the use of smaller architectures in domain-specific sub-languages and their benefits for French legal text. We prove that domain-specific pre-trained models can perform better than their equivalent generalised ones in the legal domain. Finally, we release JuriBERT, a new set of BERT models adapted to the French legal domain.
\end{abstract}

\section{Introduction}
Domain-specific language models have evolved the way we learn and use text representations in natural language processing. Instead of using general purpose pre-trained models that are highly skewed towards generic language, we can now pre-train models that better meet our needs and are highly adapted to specific domains, like medicine and law. In order to achieve that, models are trained on large scale raw text data, which is a computationally expensive step, and then are used in many downstream evaluation tasks, achieving state-of-the-art results in multiple explored domains.

The majority of domain-specific language models so far are applied to the English language. \citet{abdine-etal-2021-frenchword2vec} published French word vectors from large scale generic web content that surpassed previous non pre-trained word embeddings. Furthermore, \citet{martin-etal-2020-camembert} introduced CamemBERT, a monolingual language model for French, that is used for generic everyday text, and proved its superiority in comparison with other multilingual models. 
In the meantime, domain-specific language models for French are in lack. There is an even greater shortage when it comes to the legal field. \citet{sulea-etal-2017-legal-class} mentioned the importance of using state-of-the-art technologies to support law professionals and provide them with guidance and orientation. Given this need, we introduce JuriBERT, a new set of BERT models pre-trained on French legal text. We explore the use of smaller models architecturally when we are dealing with very specific sub-languages, like French legal text. Thus, we publicly release JuriBERT\footnote{You can find the models in \url{http://master2-bigdata.polytechnique.fr/FrenchLinguisticResources/resources\#juribert}} in 4 different sizes online.

\section{Related Work}
Previous work on domain-specific text data has indicated the importance of creating domain-specific language models. These models are either adaptations of existing generalised models, for example Bert Base by \citet{devlin-etal-2019-bert} trained on general purpose English corpora, or pre-trained from scratch on new data. In both cases, domain-specific text corpora are used to adjust the model to the peculiarities of each domain. 

A remarkable example of adapting language models is the research done by \citet{lee-etal-2019-biobert} who introduced BioBERT, a domain-specific language representation model pre-trained on large scale biomedical corpora. BioBERT outperformed BERT and other previous models on many biomedical text mining tasks and showed that pre-training on specific biomedical corpora improves performance in the field. Similar results were presented by \citet{beltagy-etal-2019-scibert} that introduced SciBERT and showed that pre-training on scientific-related corpus improves performance in multiple domains, and by \citet{yang-etal-2020-finbert} who showed that FinBERT, pre-trained on financial communication corpora, can outperform BERT on three financial sentiment classification tasks.

Moving on to the legal domain, \citet{bambroo-awasthi-2021-legalDB} worked on LegalDB, a DistilBERT model \cite{sanh-debut-etal-2019-distilbert} pre-trained on English legal-domain specific corpora. LegalDB outperformed BERT at legal document classification. \citet{elwany-etal-2019-bertlawschool} also proved that pre-training BERT can improve classification tasks in the legal domain and showed that acquiring large scale English legal corpora can provide a major advantage in legal-related tasks such as contract classification. Furthermore, \citet{chalkidis-etal-2020-legal} introduced LegalBERT, a family of English BERT models, that outperformed BERT on a variety of datasets in text classification and sequence tagging. Their work also showed that an architecturally large model may not be necessary when dealing with domain-specific sub-languages. A representative example is Legal-BERT-Small that is highly competitive with larger versions of LegalBert. We intent to further explore this theory with even smaller models. 

Despite the increasing use of domain-specific models, we have mainly been limited to the English language. On the contrary, in the French language, little work has been done on the application of text classification methods to support law professionals, with the exception of \citet{sulea-etal-2017-legal-class} that managed to achieve state-of-the-art results in three legal-domain classification tasks. It is also worth mentioning \citet{garneau-etal-2021-criminelbart} who introduced CriminelBART, a fine-tuned version of BARThez \cite{kamal-eddine-etal-2020-barthez}. CriminelBART is specialised in criminal law by using French Canadian legal judgments. All in all, no previous work has adapted a BERT model in the legal domain using French legal text.

\section{Downstream Evaluation Tasks}
In order to evaluate our models we will be using two legal text classification tasks provided by the Court of Cassation, the highest court of the French judicial order. 

The subject of the first task is assigning the Court’s Claimant’s pleadings, "m\`emoires ampliatifs" in French, to a chamber and a section of the Court. This leads to a multi-class classification task with 8 different imbalanced classes. In Table \ref{tab:court_classes} we can see the eight classes that correspond to the different chambers and sections of the Court, as well as their support in the data. The classes represent 4 chambers: the first civil chamber (C1) that deals with topics like Civil Contract Law and Consumer Law, the second civil chamber (C2) with topics like Insurance Law and Traffic accidents, the third civil chamber (C3) dealing with Real property and Construction Law among other topics and the Commercial, Economic and Financial Chamber (CO) for Commercial Law, Banking and Credit Law and others. Each chamber has two or more sections dealing with different topics.

\begin{figure}
\centering
\includegraphics[width=0.45\textwidth]{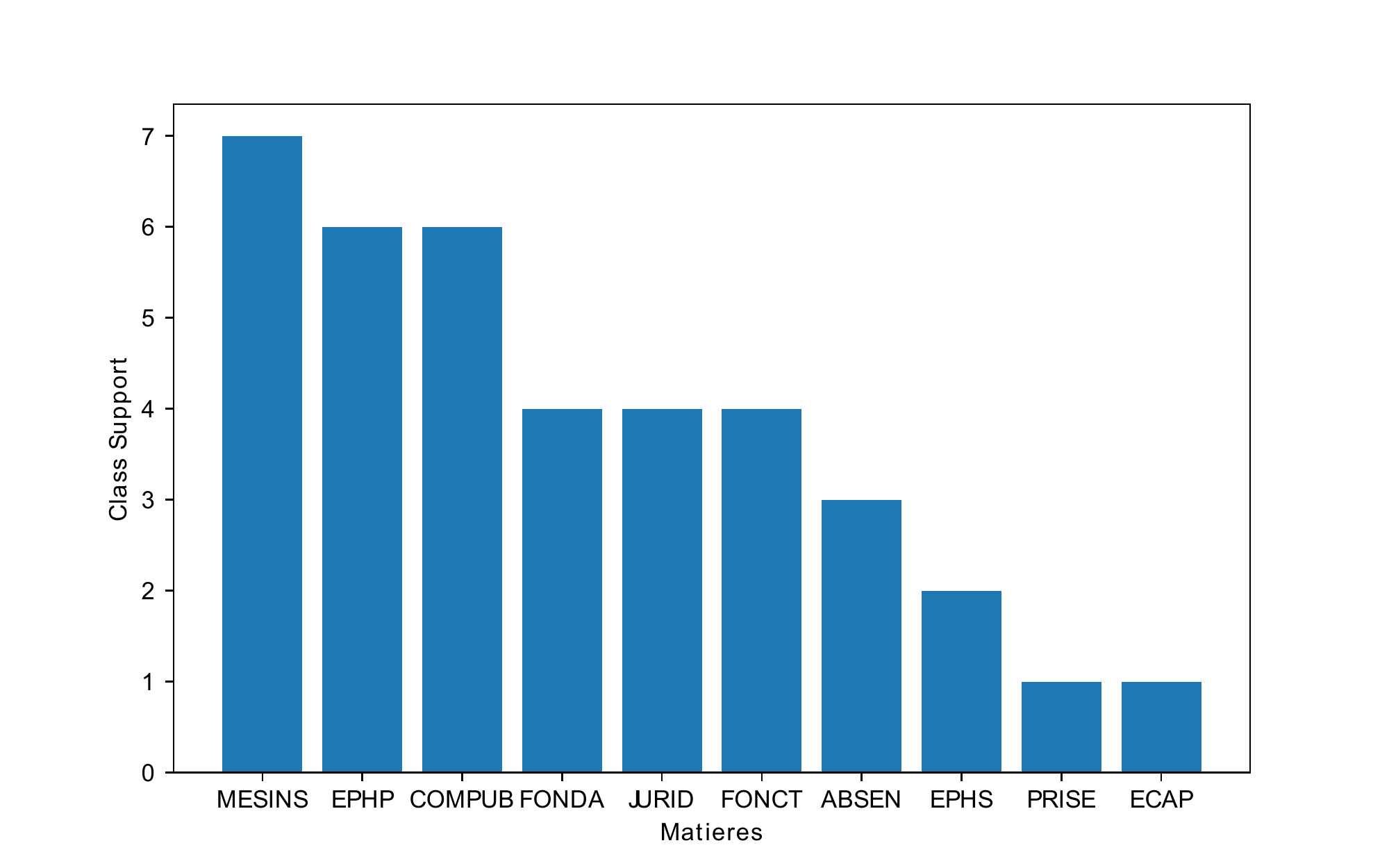}
\caption{10 Recessive Matieres}
\label{fig:recessive_matieres}
\end{figure}

The second task is to classify the Claiment's pleadings to a set of 151 subjects, "mati\`eres" as stated in French. Figure \ref{fig:matieres_support} in appendix shows the support of the mati\`eres in the data. As we can see in Figure \ref{fig:recessive_matieres} the 10 recessive mati\`eres have between 7 to 1 examples in our dataset. We decided to remove the last 3 mati\`eres as they have less than 3 examples and therefore it is not possible to split them in train, test and development sets.

\begin{table}
\centering
\begin{tabular}{lc}
\hline
\textbf{Class} & \textbf{Support} \\
\hline
CO &  28 198 \\
C1\_Section1 &  14 650 \\
C1\_Section2 & 16 730 \\
C2\_Section1 &  11 525 \\
C2\_Section2 &  9 975 \\
C2\_Section3 & 13 736 \\
C3\_Section1 &  16 176 \\
C3\_Section2 & 12 282 \\ \hline
\end{tabular}
\caption{Chambers and Sections of the Court of Cassation and data support}
\label{tab:court_classes}
\end{table}

\section{JuriBERT}
We introduce a new set of BERT models pre-trained from scratch in legal-domain specific corpora. We train our models on the Masked Language Modeling (MLM) task. This means that given an input text sequence we mask tokens with 15\% probability and the model is then trained to predict these masked tokens. We follow the example of \citet{chalkidis-etal-2020-legal} and choose to train significantly even smaller models, including Bert-Tiny and Bert-Mini. The architectural details of the models we pre-trained are presented in Table \ref{tab:architectures}.
We also choose to further pre-train CamemBERT Base on French legal text in order to better explore the impact of using domain-specific corpora in pre-training.

\subsection*{Training Data}
For the pre-training we used two different French legal text datasets. The first dataset contains data crawled\footnote{We used Heritrix, a crawler that respects the robots.txt exclusion directives and META nofollow tags. See \url{https://github.com/internetarchive/heritrix3}} from the L\'egifrance\footnote{\url{https://www.legifrance.gouv.fr/}} website and consists of raw French Legal text. The L\'egifrance text is then cleaned from non French characters. We also use the Court's decisions and the Claimant's pleadings from the Court of Cassation that consists of 123361 long documents from different court cases. All personal and private information, including names and organizations, has been removed from the documents for the privacy of the stakeholders. The combined datasets provide us with a collection of raw French legal text of size 6.3 GB that we will use to pre-train our models.

\subsection*{Legal Tokenizer}
In order to pre-train a new BERT model from scratch we need a new Tokenizer. We trained a ByteLevelBPE Tokenizer with newly created vocabulary from the training corpus. The vocabulary is restricted to 32,000 tokens in order to be comparable to the CamemBERT model from \citet{martin-etal-2020-camembert} and minimum token frequency of 2.  We used a RobertaTokenizer as a template to include all the necessary special tokens for a Masked Language Model. Our new Legal Tokenizer encodes the data using 512-sized embeddings.

\subsection*{JuriBERT}
For the pre-training of the JuriBERT Model we used both the crawled L\'egifrance data and the Pleadings Dataset, thus creating a 6.3GB collection of legal texts. The encoded corpus was then used to pre-train a BERT model from scratch. Our model was pre-trained in 4 different architectures. As a result we have JuriBERT Tiny with 2 layers, 128 hidden units and 2 attention heads (6M parameters), JuriBERT Mini with 4 layers, 256 hidden units and 4 attention heads (15M parameters), JuriBERT Small with 6 layers, 512 hidden units and 8 attention heads (42M parameters) and JuriBERT Base with 12 layers, 768 hidden units and 12 attention heads (110M parameters). JuriBERT Base uses the exact same architecture as CamemBERT Base.

\subsection*{Task Specific JuriBERT}
We also pre-trained a task-specific model that is expected to perform better in the classification of the Claiment's pleadings. For that we used only the Pleadings Dataset for the pre-training that is a 4GB collection of legal documents. The task-specific JuriBERT model for the Cour de Cassation task was pre-trained in 2 architectures, JuriBERT Tiny (L=2, H=128, A=2) and JuriBERT Mini (L=4, H=256, A=4).

\subsection*{JuriBERT-FP}
Apart from pre-training from scratch we decided to also further pre-train CamemBERT Base on the training data. Our goal is to compare its performance with the original JuriBERT model to further explore the impact of using specific-domain corpora during pre-training.
JuriBERT-FP uses the same architecture as CamemBERT Base and JuriBERT Base.

\section{Methods}
\paragraph{Pre-training Details}
All the models were pre-trained for 1M steps. A learning rate of $1e-4$ was used along with an Adam optimizer ($\beta_1$=0.9, $\beta_2$=0.999) with weight decay of 0.1 and a linear scheduler with 10,000 warm-up steps. All the models were pre-trained with batch size of 8 expect for JuriBERT Base and JuriBERT-FP that used batches of size 4. For the pre-training we used an Nvidia GTX 1080Ti GPU.

\begin{table}
\centering
\begin{tabular}{lcc}
\hline
\textbf{Model} & \textbf{Architecture} & \textbf{Params}\\
\hline
JuriBERT Tiny & L=2, H=128, A=2 & 6M \\
JuriBERT Mini & L=4, H=256, A=4 & 15M \\
JuriBERT Small & L=6, H=512, A=8 & 42M \\
JuriBERT Base & L=12, H=768, A=12 & 110M \\
JuriBERT-FP & L=12, H=768, A=12 & 110M
\\\hline
\end{tabular}
\caption{Architectural comparison of JuriBERT models}
\label{tab:architectures}
\end{table}

\begin{table}
\centering
\begin{tabular}{lc}
\hline
Model & Pre-training Corpora \\
\hline
CamemBERT & 138GB  \\
BARThez & 66GB \\
JuriBERT & 6.3GB \\
JuriBERT-FP & 6.3GB \\
Task JuriBERT & 4GB \\\hline
\end{tabular}
\caption{Size of pre-training corpora used by different models}
\label{tab:corpora_size}
\end{table}

\paragraph{Fine-tuning Details}
Our models were fine-tuned on the downstream evaluation task using the same classification head as \citet{devlin-etal-2019-bert} that consists of a Dense layer with tanh function followed by a Dense layer with softmax activation function and Dropout layers with fixed dropout rate of 0.1. We applied grid-search to the learning rate on a range of \{$2e-5, 3e-5, 4e-5, 5e-5$\}. We used an Adam optimizer along with a linear scheduler that provided the training with 100k warm-up steps. We train for a maximum of 30 epochs with patience of 2 epochs on the early stopping callback and checkpoints for the best model. For the classification we use only the paragraphs starting with 'ALORS QUE' from the Pleadings Dataset, as they include all the important information for the correct chamber and section. This was suggested by a lawyer from the Court of Cassation as the average size of a m\`emoire ampliatif is extremely big, from 10 to 30 pages long. By using the 'ALORS QUE' paragraphs we have text sequences with average size of 800 tokens. For the chambers and sections classification task we split the data in 14\% development and 16\% test data. For the mati\`eres classification we split the data in 17\% development and 14\% test data and stratify in order to have all classes represented in each subset. Both tasks use a fixed batch size of 4. For the fine-tuning we used an Nvidia GTX 1080Ti GPU.

\section{Results}

The results on the downstream evaluation tasks are presented in Tables \ref{tab:accuracy} and \ref{tab:accuracy2}.
We compare our models with two CamemBERT versions, Base and Large, and with BARThez, a sequence-to-sequence model dedicated to the French language. CamemBERT has been pre-trained on 138GB of French raw text from the OSCAR corpus. Despite the difference in pre-training corpora size, with our model using only 6.3GB of legal text, JuriBERT Small managed to outperform both CamemBERT Base and CamemBERT Large. This further proves the importance of domain-specific language models in natural language processing and transfer learning. Despite our expectations, the performance of JuriBERT Base does not exceed the performance of its smaller equivalent models. We attribute this peculiarity in the usage of smaller batch sizes when pre-training JuriBERT Base and also the fact that larger models usually need more computational resources and more time and data in order to converge. 

JuriBERT Small also outperforms BARThez on the chambers and sections evaluation task, which is pre-trained on 66GB of French raw text and usually used for generative tasks. On the mati\`eres classification task BARThez is the dominant model with JuriBERT Small being second. We infer that the complexity of the second task benefits more from the robustness and size of BARThez than from the specific-domain nature of JuriBERT.

Comparing our models with the same architectures, it becomes apparent that all task-specific JuriBERT models perform better than their equivalent domain-specific JuriBERT models besides using only 4GB of pre-training data. The results confirm that a BERT model pre-trained from scratch only on the corpus that is then used for fine-tuning can perform better than a domain-specific one on the same task as we expected.

JuriBERT-FP outperforms JuriBERT Base and achieves similar results with CamemBERT Base on the chambers and sections classification task. This shows that further pre-training a general purpose language model can have better results than training from scratch. However, it did not manage to outperform JuriBERT Small in both tasks, which can be attributed to the smaller batch size used during pre-training and to the size of the model as mentioned before for JuriBERT Base. Unfortunately, there are no smaller versions of CamemBERT available to further test this theory. On the mati\`eres classification task, JuriBERT-FP still outperforms JuriBERT Base. On the contrary, it performs worse than CamemBERT Base. Along with the state-of-the-art results of BARThez, this leads us to believe that in order to achieve better results in more complex tasks JuriBERT models require more pre-training corpora.

All in all, JuriBERT Small achieves equivalent results with previous larger generic language models with an accuracy of 83.95\% on the first task and 71.80\% on the second task on the test data. JuriBERT Small, JuriBERT Mini and even JuriBERT Tiny all outperform JuriBERT Base, proving that smaller models architecturally can achieve comparable, if not better, results when we are training on very domain-specific data. A larger model, not only requires more resources to be trained, but is also not as efficient as its smaller equivalents. This is of major importance for researchers with limited resources available. Furthermore, JuriBERT-FP achieves better results than JuriBERT Base in both tasks. This leads us to infer that pre-training from an existing language model can be a major advantage, as opposed to randomly initialising our weights.

\begin{table}
\centering
\begin{tabular}{lccc}
\hline
\textbf{Model} & \textbf{Lrate} & \textbf{Dev} & \textbf{Test}\\
\hline
CamemBERT Base & $2e-5$ & 82.75 & 83.22 \\
CamemBERT Large & $2e-5$ & 79.69 & 79.91 \\
\hline
BARThez & $3e-5$ & 83.70 & 83.49 \\
\hline
JuriBERT Tiny & $3e-5$ & 82.00 & 81.58 \\
JuriBERT Mini & $3e-5$ & 83.08 & 82.62 \\
JuriBERT Small & $3e-5$ & 83.86 & \textbf{83.95} \\
JuriBERT Base & $3e-5$ & 82.26 & 82.51 \\ 
\hline
Task JuriBERT Tiny & $4e-5$ & 81.91 & 81.59 \\
Task JuriBERT Mini & $4e-5$ & 82.75 & \textbf{82.66} \\
\hline
JuriBERT-FP & $2e-5$ & 83.07 & \textbf{83.28} \\
\hline
\end{tabular}
\caption{Accuracy of models on the chambers and sections classification task}
\label{tab:accuracy}
\end{table}

\begin{table}
\centering
\begin{tabular}{lccc}
\hline
\textbf{Model} & \textbf{Lrate} & \textbf{Dev} & \textbf{Test}\\
\hline
CamemBERT Base & $3e-5$ & 71.64 & 71.66 \\
\hline
BARThez & $2e-5$ & 72.17 & \textbf{72.09} \\
\hline
JuriBERT Small & $2e-5$ & 71.67 & \textbf{71.80} \\
JuriBERT Base & $3e-5$ & 70.28 & 70.38 \\ 
\hline
JuriBERT-FP & $2e-5$ & 70.99 & \textbf{71.21} \\
\hline
\end{tabular}
\caption{Accuracy of models on the matieres classification task}
\label{tab:accuracy2}
\end{table}




\section{Limitations}
As we mentioned before both JuriBERT Base and JuriBERT-FP have been pre-trained using smaller batch sizes than the other models due to limited resources. We acknowledge that this may have affected their performance compared to the other models. However, we believe that their lower performance can also be attributed to their size as larger models are computationally heavier and thus require more resources to converge. 

Acquiring large scale legal corpora, especially for a language other than English, has proven to be challenging due to their confidential nature. For this reason, JuriBERT models were fine-tuned on two downstream evaluation tasks that contain data from the pre-training dataset collection.  Further testing shall be required in order to validate the performance of our models on different tasks.

The differences in performance between the generic language models and the newly created JuriBERT models are very small. More specifically, only JuriBERT Small manages to outperform CamemBERT Base and Barthez with a difference in accuracy of 0.73\%. We attribute this limitation in the use of much less pre-training data. However we emphasize that JuriBERT manages to achieve similar results despite the difference in pre-training corpora size. Thus, we expect JuriBERT to achieve better results in the future provided that we further pre-train with more data.

\section{Conclusions and Future Work}
We introduce a new set of domain-specific Bert Models pre-trained from scratch on French legal text. We conclude that our task is very specific and as a result it does not benefit from general purpose models like CamemBERT. We also show the superiority of much smaller models when training on very specific sub-languages like legal text. It becomes apparent that large architectures may in fact not be necessary when the targeted sub-language is very specific. This is important for researchers with lower resources available, as smaller models are fine-tuned a lot faster on the downstream tasks. Furthermore, we show that a BERT model pre-trained from scratch on task-specific data and then fine-tuned on this very task can perform better than a domain-specific model that has been pre-trained on a lot more data. We point out of course that a domain-specific model can outperform a task-specific one on other tasks and is generally preferred when we need a multi-purpose BERT model with many applications in the French legal domain. In future work, we plan to further explore the potential of JuriBERT in other tasks and as a result prove its superiority over the task-specific one.

\section*{Acknowledgements} 
This research was supported by the ANR chair AML/HELAS (ANR-CHIA-0020-01).

\bibliography{main}
\bibliographystyle{acl_natbib}

\clearpage

\appendix
\section{Appendix}
\label{sec:appendix}


\begin{figure}[ht]
\centering
\includegraphics[width=0.45\textwidth]{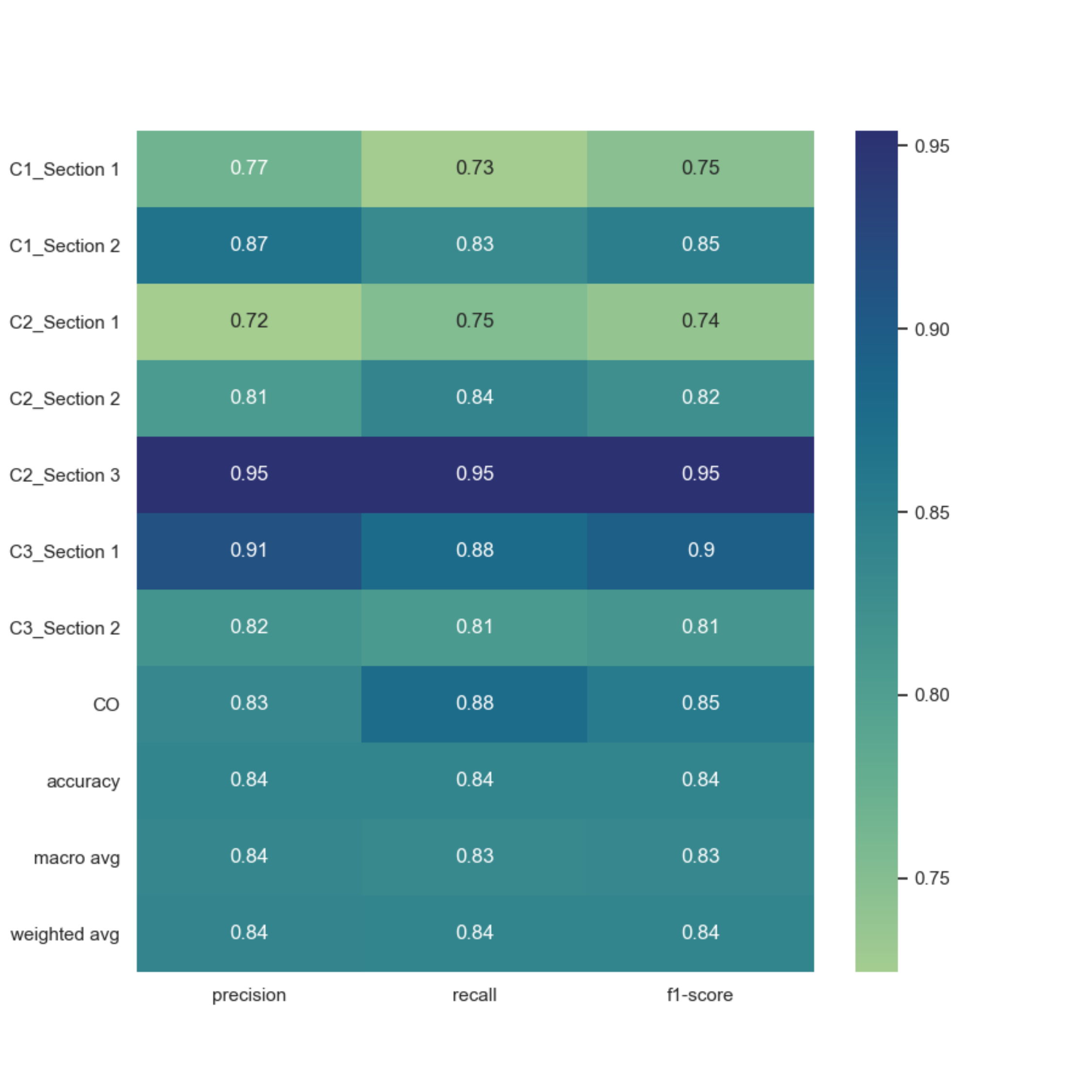}
\caption{Accuracy, Precision, Recall and F1-Score of JuriBERT Small on the chambers and sections classification task on the test dataset. The graph contains all 8 classes.}
\label{fig:class_report_chambers}
\end{figure}

\begin{figure}[ht]
\centering
\includegraphics[width=0.45\textwidth]{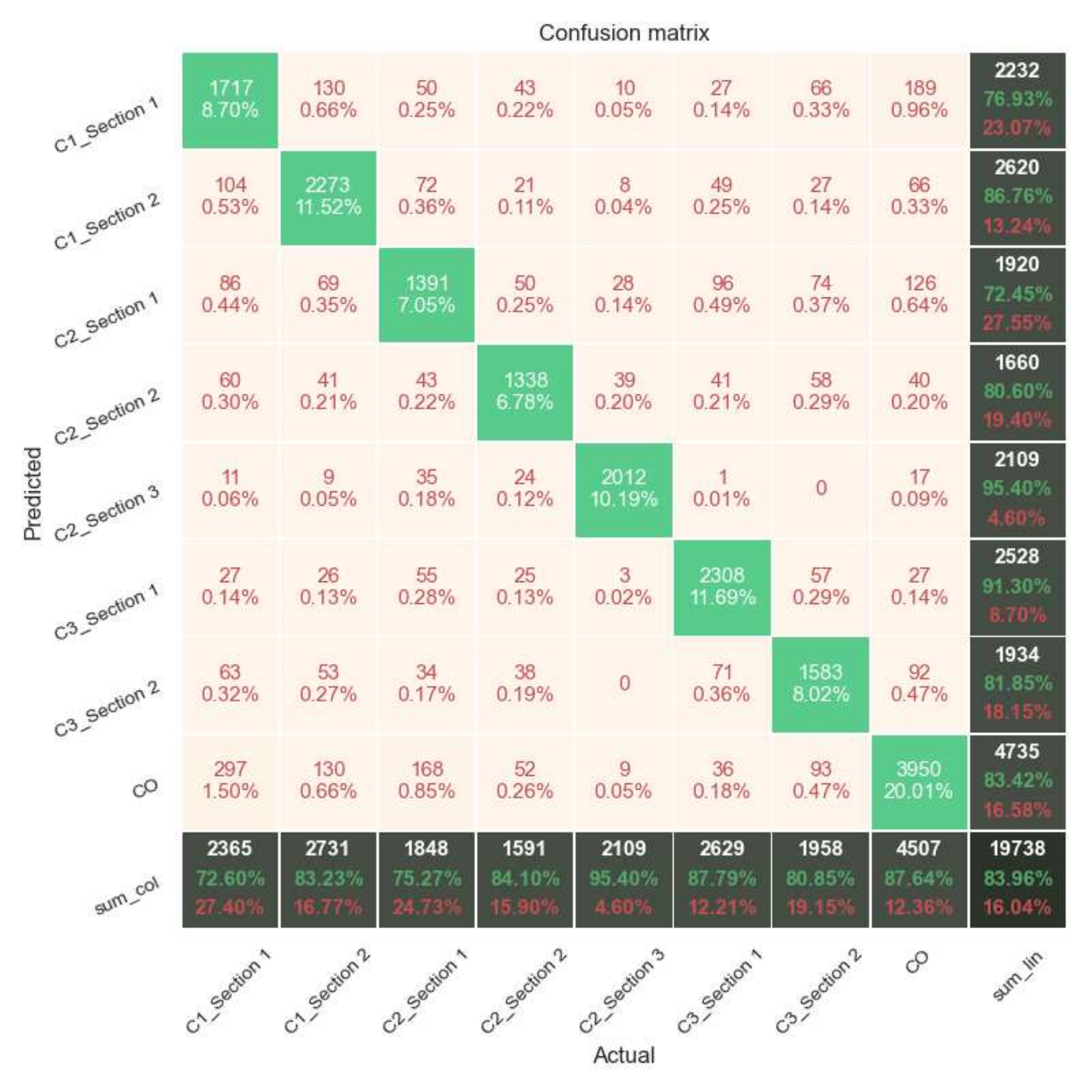}
\caption{Confusion Matrix of JuriBERT Small on the chambers and sections classification task on the test dataset. The graph includes accuracy and error rate for each class.}
\label{fig:confusion_matrix_chambers}
\end{figure}

\begin{figure}[ht]
\centering
\includegraphics[width=0.45\textwidth]{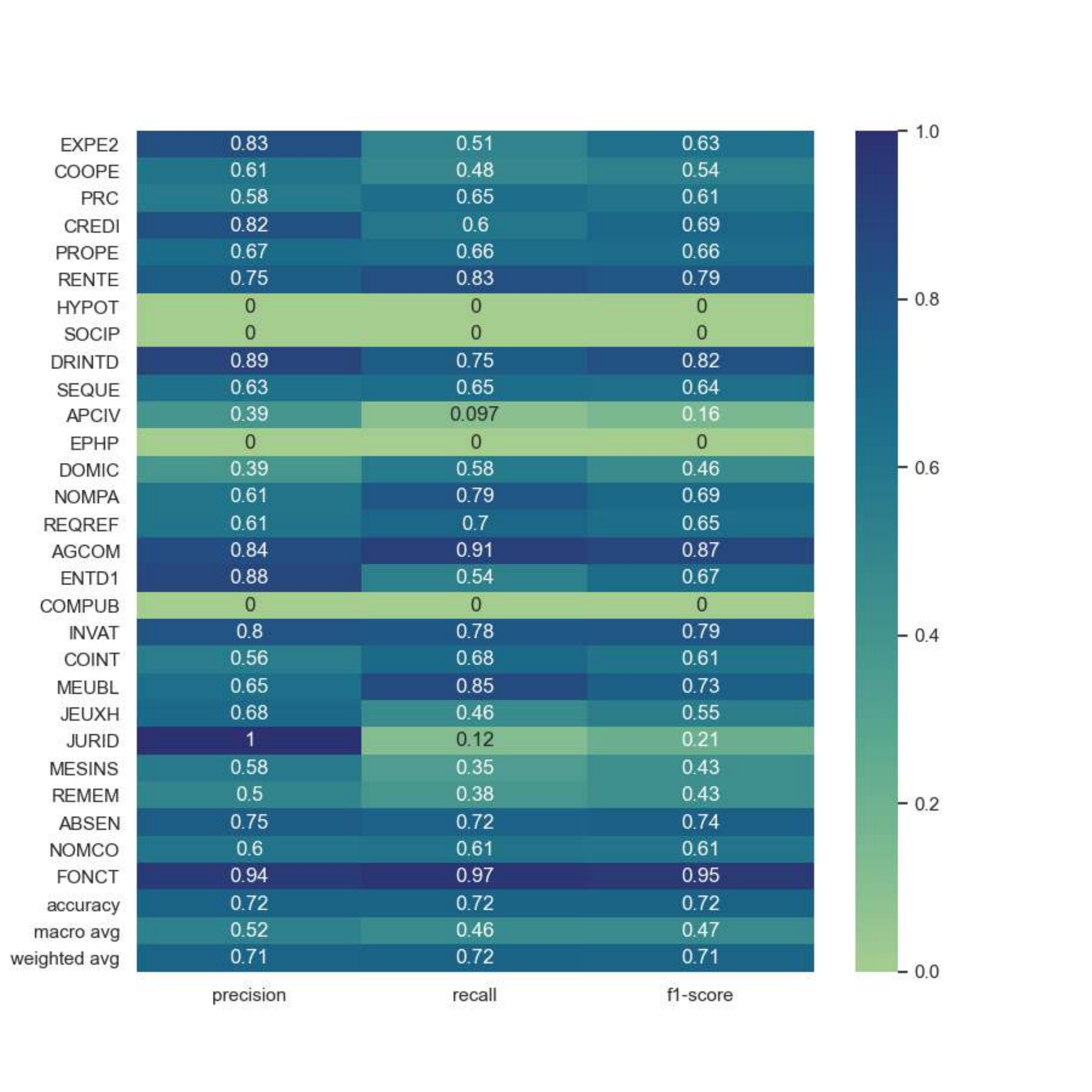}
\caption{Sample of Accuracy, Precision, Recall and F1-Score of JuriBERT Small on the matieres classification task on the test dataset. The graph contains 28 classes and the overall accuracy of all 148 classes.}
\label{fig:class_report_matieres}
\end{figure}

\begin{figure}[ht]
\centering
\includegraphics[width=0.4\textwidth]{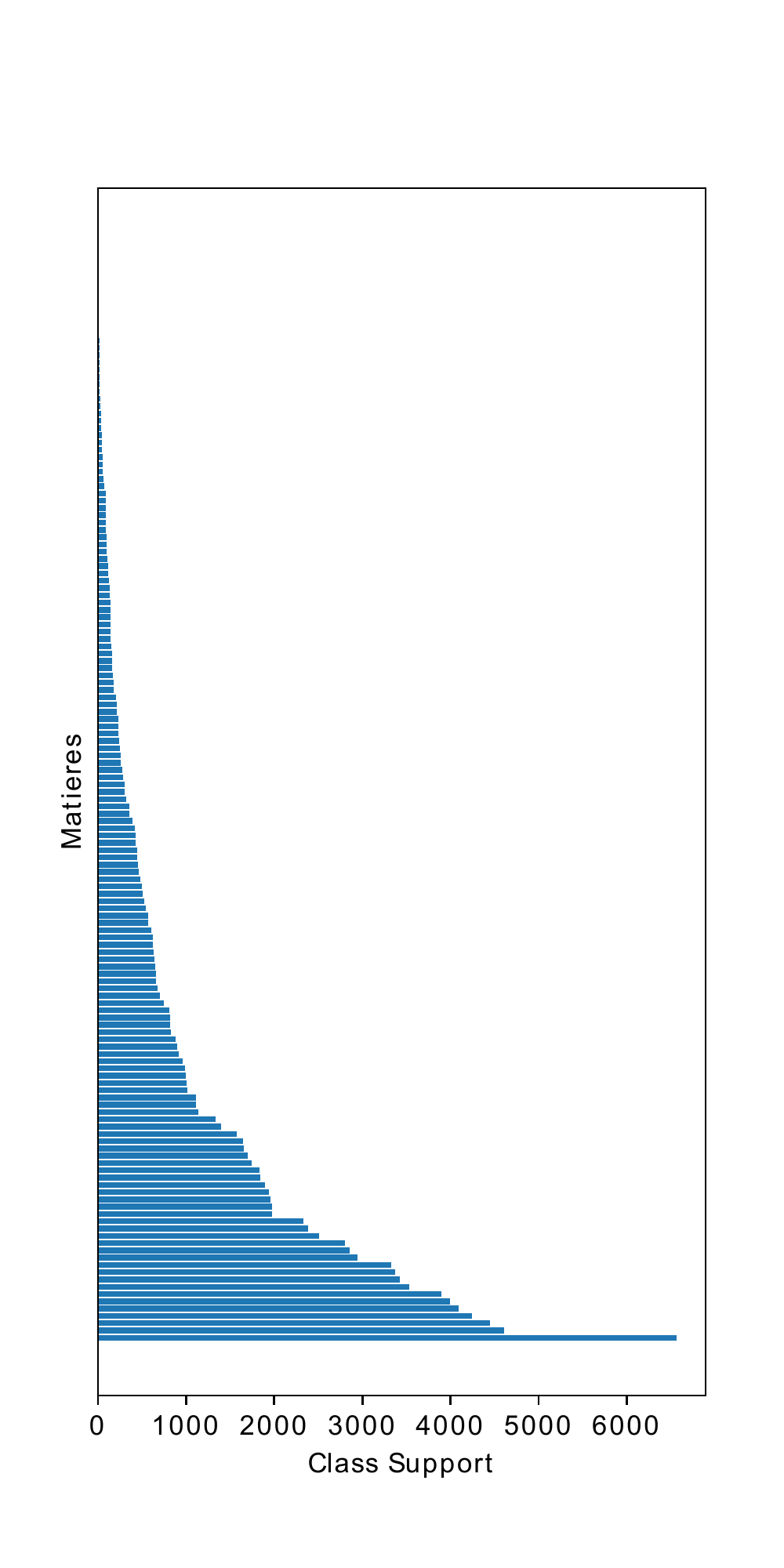}
\caption{Support of the 151 Matieres in the Court of Cassation data.}
\label{fig:matieres_support}
\end{figure}

\end{document}